\documentclass[times, 10pt,twocolumn]{article}
\usepackage{latex12}
\usepackage{times}
\usepackage{balance}
\usepackage{amsmath, graphicx, xspace, amsfonts}
\usepackage{enumitem}
\setlist{nolistsep}
\usepackage{tikz}
\usepackage{url}

\graphicspath{{figures/}}

\begin{document}

\title{Morphological Filtering in Shape Spaces :\\ Applications using Tree-Based Image Representations}

\author{Yongchao Xu$^{1,2}$ and Thierry G\'eraud$^{1}$\\
\emph{$^{1}$ EPITA Research and Development Laboratory (LRDE)}\\
\emph{\{yongchao.xu,thierry.geraud\}@lrde.epita.fr}\\
%
%
\and
Laurent Najman$^{2}$\\
\emph{$^{2}$ Universit\'e Paris-Est, LIGM, \'Equipe A3SI, ESIEE}\\
\emph{l.najman@esiee.fr}
}

\maketitle
\thispagestyle{empty}

\long\def\symbolfootnote[#1]#2{\begingroup%
\def\thefootnote{$\dag$}\footnote[#1]{#2}\endgroup}

\begin{abstract}
  Connected operators are filtering tools that act by merging
  elementary regions of an image. A popular strategy is based on
  tree-based image representations: for example, one can compute an
  attribute on each node of the tree and keep only the nodes for which
  the attribute is sufficiently strong. This operation can be seen as
  a thresholding of the tree, seen as a graph whose nodes are weighted
  by the attribute. Rather than being satisfied with a mere
  thresholding, we propose to expand on this idea, and to apply
  connected filters on this latest graph. Consequently, the filtering is
  done not in the space of the image, but on the space of shapes build
  from the image.

  Such a processing is a generalization of the existing tree-based
  connected operators. Indeed, the framework includes classical
  existing connected operators by attributes.
  It also allows us to propose a class of novel connected operators
  from the leveling family, based on shape attributes. Finally, we
  also propose a novel class of self-dual connected operators that we
  call {\em morphological shapings}.
\end{abstract}

\newcommand{\Chi}{\ensuremath{\mathcal{X}}\xspace}
\newcommand{\Attribute}{\ensuremath{\mathcal{A}}\xspace}
\newcommand{\Tree}{\ensuremath{\mathcal{T}}\xspace}

\newcommand{\thickness}{\ensuremath{\varepsilon}\xspace}
\newcommand{\curv}{\ensuremath{\mathit{curv}}\xspace}

\newcommand{\Rin}{\ensuremath{\mathcal{R}_{in}^\thickness(\partial\tau)}\xspace}
\newcommand{\Rout}{\ensuremath{\mathcal{R}_{out}^\thickness(\partial\tau)}\xspace}

\newcommand{\fdt}{\ensuremath{f, \partial\tau}\xspace}

\newcommand{\Eint}{\ensuremath{E_\mathit{int}}\xspace}
\newcommand{\Eext}{\ensuremath{E_\mathit{ext}}\xspace}
\newcommand{\Econ}{\ensuremath{E_\mathit{con}}\xspace}

\newcommand{\snake}{\ensuremath{^\mathit{snk}}}
\newcommand{\Eintsnk}{\ensuremath{\Eint\snake}\xspace}
\newcommand{\Eextsnk}{\ensuremath{\Eext\snake}\xspace}

\newcommand{\Elocal}{\ensuremath{E^\mathit{local}}\xspace}
\newcommand{\Eglobal}{\ensuremath{E^\mathit{global}}\xspace}

\newcommand{\setRi}{\ensuremath{\{\mathcal{R}_i\}}\xspace}
\newcommand{\settau}{\ensuremath{\{\tau\}}\xspace}

\newcommand*\mycirc[1]{%
  \begin{tikzpicture}
    \node[draw,circle,inner sep=0.2pt] {#1};
  \end{tikzpicture}}

\newcommand{\notsoclose}{2mm}
\newcommand{\closer}{-4\mm}

\Section{Introduction}
\label{sec:intro}

\begin{figure}
  \begin{center}
  \begin{minipage}[b]{0.8\linewidth}
    \includegraphics[width=1.0\linewidth]{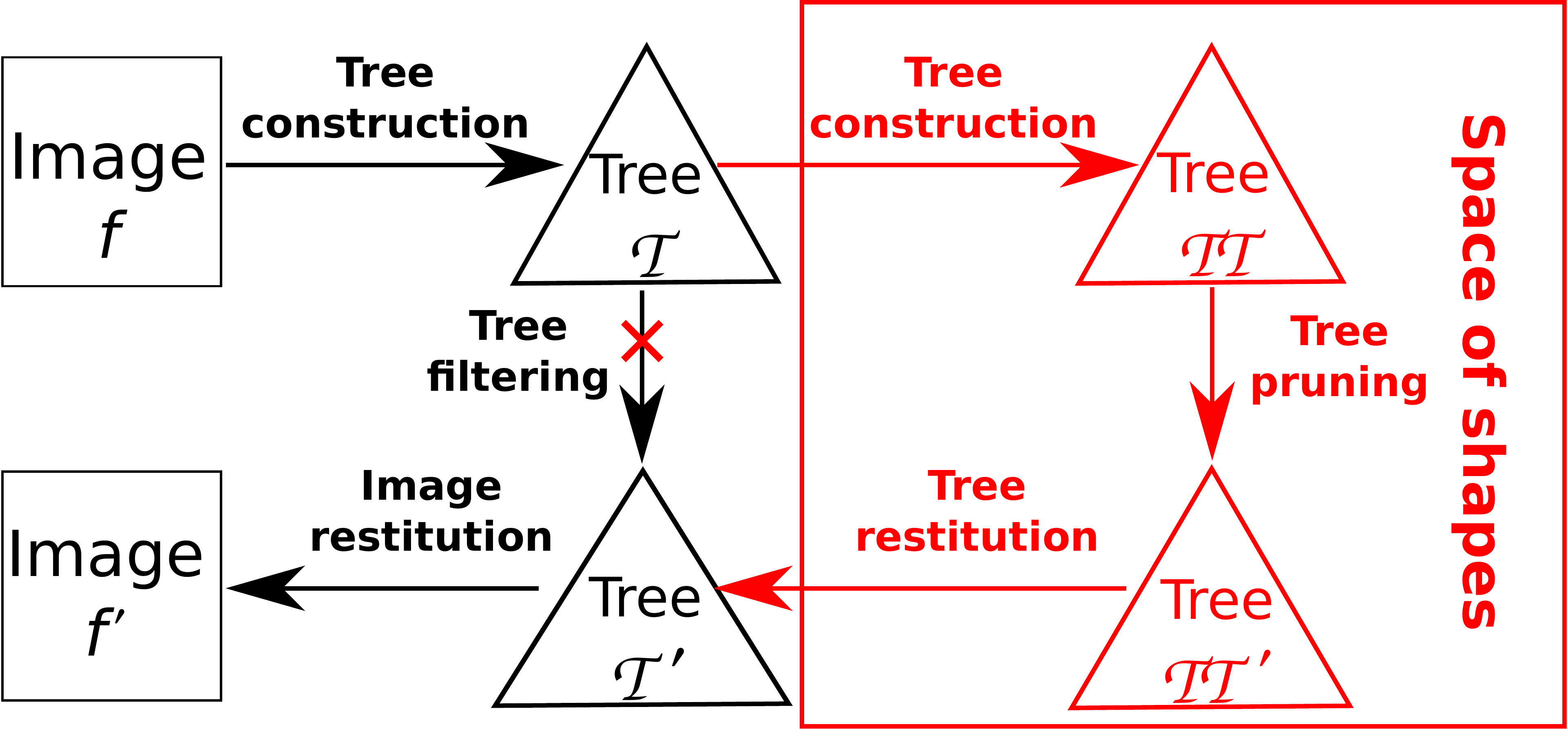}
  \end{minipage}
  \end{center}
  \vspace*{-5mm}
  \caption{Classical connected operators (black path) and our proposed
    process (by adding the red path).}
  \label{fig:shema}
\end{figure}

Mathematical morphology, as originally developed by Matheron and
Serra~\cite{serra.82.ap}, proposes a set of morphological operators based
on structuring elements. Later, Salembier and Serra
\cite{salembier.95.itip}, followed by Breen and Jones
\cite{breen.96.cviu}, proposed morphological operators based on
attributes, rather than on elements. Such operators rely on
transforming the image into an equivalent representation, generally a
tree of components ({\em e.g.}  level sets) of the image; such trees
are equivalent to the original image in the sense that the image can
be reconstructed from its associated tree. Filtering then involves the
design of a shape attribute that weights how much a node of the tree
fits a given shape.  Two different approaches for filtering the tree
(and hence the image) have been proposed: the more evolved approach
consists in pruning the tree by removing whole branches of the tree,
and is easy to apply if the attribute is increasing on the tree ({\em
  i.e.}, if the attribute is always stronger for the ancestors of a
node).  The process is illustrated in Figure~\ref{fig:shema} by the
black path.

However, most shape attributes are not increasing. When the attribute
is not increasing, three strategies have been proposed (min, max,
Viterbi; see \cite{salembier.09.spm} for more details). They all
choose a particular node on which to take the decision, and remove the
whole subtree rooted in this node.  While it may give interesting
results in some cases, it does not take into account the possibility
that several relevant objects can have some inclusion relationship, which means that they are on the same branch of the tree
(for instance a ring object in a tree of shapes, see
Figure~\ref{fig:comparison}~(a)).

In the simplest approach, one simply removes the nodes of the tree for
which the attribute is lower than a given
threshold~\cite{urbach.07.pami}.  Such a thresholding does not take
into account the intrinsic parenthood relationship of the tree.
Moreover it is often impossible to retrieve all expected objects with
one unique threshold.

The founding idea of this paper is to apply connected filters on the
space made by all the components of the image, such space being
structured into a graph by the parenthood relationship (i.e., a node
has for neighbors its children and its parent).  This is the red path
in Figure~\ref{fig:shema}. This surprising and simple idea has several
deep consequences, that form the main contributions of this paper.
First, we show that the framework encompasses the usual attribute
filtering operators.  Second, novel connected filters based on
non-increasing criterion can be proposed, and we show that such
filters are new morphological levelings~\cite{meyer.04.jmiv}.  And
third, we propose a novel family of self-dual connected filters that
we call {\em morphological shapings}.

The rest of this paper is organized as follows.  Our proposed
shape-based morphology is explained in
Section~\ref{sec:shape.based.morpho}.  In Section~\ref{sec:results},
we depict some experimental results.  Finally we conclude in
Section~\ref{sec:conclusion}.

\Section{Shape-based morphology}
\label{sec:shape.based.morpho}


As stated in the introduction, one way to compute connected operators
is to represent the input image $f$ by a component tree $\Tree$,
either a min-tree (given by the inclusion relationship of the
connected components of the lower level sets), a
max-tree~\cite{salembier.98.itip} (upper level sets), or a tree of
shapes~\cite{monasse.00.itip}. The connected components are given
thanks to a neighborhood, usually C4 or C8. Let us remark that the
tree $\Tree$ with its nodes valued with an attribute $\Attribute$ can
be seen as a node-weighted graph where vertices are tree nodes,
adjacency (graph edges) is parenthood.
Such a graph is a shape-space, as any node is a component of the
original image. As an edge of the graph represents an inclusion
relationship between components, the neighborhood of a node is a kind
of ``context'' of the node (component) in the image.

If, for example, $\Attribute$ encodes the probability for a component
to be of a given type, the minima of the space of shapes are the
components that are the less probable to be of that type, compared to
their parents and children.  As a node-weighted graph, the space
$\Tree$ can be represented by a component tree $\Tree\Tree$
(explicitly and for clarity a min-tree in the sequel of this
paper). Minima of $\Attribute$ on $\Tree$ are leaves of $\Tree\Tree$.
Pruning or removing some branches of the new tree $\Tree\Tree$ thus
removes the parts of $\Tree$ that are less likely to be of the type
favored by $\Attribute$.  Remark that with this strategy, we can
remove two different ``objects'' located in the same branch of
$\Tree$, but represented by two different branches of $\Tree\Tree$.

Let us now compare our approach with the state of the art in connected
filtering. The regular case is when the attribute $\Attribute$ is
increasing, {\em i.e.} when $\Attribute$ is lower for a component than
for its parent. In that case, the min-tree $\Tree\Tree$ is equal to
$\Tree$. Taking the red path in Figure~\ref{fig:shema} is equivalent
to taking the black path. In other words, the proposed approach
encompasses the classical one.

\begin{figure}
  \begin{center}
    \begin{minipage}[b]{0.7\linewidth}
      \centerline{\includegraphics[width=0.6\linewidth]{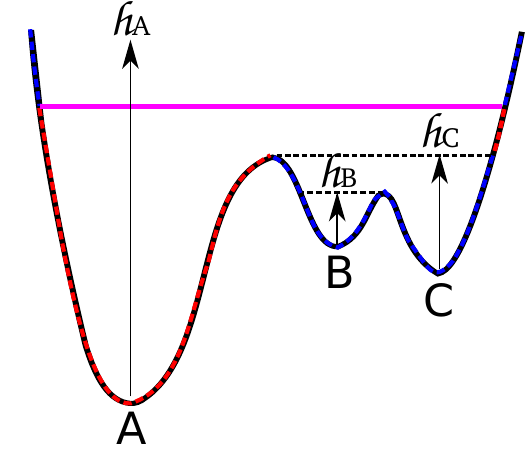}}
    \end{minipage}
  \end{center}
  \vspace*{-5mm}
\caption{Illustration of the extinction values of three minima.}
\label{fig:extinction}
\end{figure}

A shape-attribute $\Attribute$ is more often non-increasing. In such a
case, $\Tree\Tree$ is different from $\Tree$.  Figure
\ref{fig:extinction} shows an example of a non-increasing attribute in
a branch of $\Tree$.
Pruning $\Tree\Tree$ is equivalent to the thresholding of $\Attribute$
on $\Tree$. In Figure \ref{fig:extinction}, such a pruning removes the
nodes whose $\Attribute$ is under the purple line.

Another classical strategy is to compute an attribute
$\Attribute\Attribute$ on $\Tree\Tree$, and to prune the tree
according to this attribute in shape-space. In that case,
\begin{itemize}
\item when the tree $\Tree$ is a min-tree or a max-tree, such a
  morphological filtering in shape space is a
  leveling of the original image (recall that an
  operator is a leveling if and only if it preserves the order $\leq$
  and $\geq$ between neighboring pixels). We call such a filtering
  {\em shape-based leveling}.
\item when the tree $\Tree$ is a tree of shape, then the order between
  neighboring pixels is no longer preserved by the filtering, and it
  is thus no longer a leveling.  We call this new family of self-dual
  morphological connected filters the {\em{morphological shapings}}.
\end{itemize}

Of course, $\Attribute\Attribute$ can be as simple as the height of a
component in the shape space. 
But other attributes $\Attribute\Attribute$ can be computed based on
$\Attribute$ or even the image domain: for example, the area of a
connected component of $\Tree\Tree$ can be the number of pixels of this
set of components in the original image.

Let us mention another interesting variant for the filtering
strategy. It is based on the use of extinction
value~\cite{vachier.95.nlsp} of minima. Let $\prec$ be a strict total
order on the set of minima $m_1\prec m_2 \prec\ldots$, such that
$m_i\prec m_{i+1}$ whenever the altitude of $m_i$ is lower than the
altitude of $m_{i+1}$. Let $CC$ be the lowest connected component that
contains both $m_{i+1}$ and a minimum $m_j$ with $j<(i+1)$. The
extinction value for the minimum $m_{i+1}$ is defined as the
difference between the altitude of $CC$ and the altitude of $m_{i+1}$.
Figure \ref{fig:extinction} shows an example of the extinction value
for three minima. The order is $A \prec C \prec B$.  The filtering
strategy is to preserve (or remove) only the blobs determined by a
minimum whose extinction value is higher than a given value. The
advantage of this strategy is that it preserves only those shapes
which are meaningful enough compared to their context. For example,
the red part in Figure \ref{fig:extinction} is preserved for the value
given by the purple line, but the blobs corresponding to minima $B$
and $C$ are removed.


\Section{Some illustrations}
\label{sec:results}
In this section, we present some illustrations of our new filters,
both for image filtering (section~\ref{sec:objectfiltering}) and for
object detection (section~\ref{sec:segmentation}).
\SubSection{Object filtering}
\label{sec:objectfiltering}

\newcommand{\tinympscale}{0.45} 
\newcommand{\tinyfigscale}{0.9} 
\newcommand{\circularity}{$\bigcirc$}
\newcommand{\tinycloser}{-0mm}


\begin{figure}
  \begin{center}
    %
    %
    ~\hspace*{0mm}
    \begin{minipage}[b]{\tinympscale\linewidth}
    \centerline{\includegraphics[width=\tinyfigscale\linewidth]{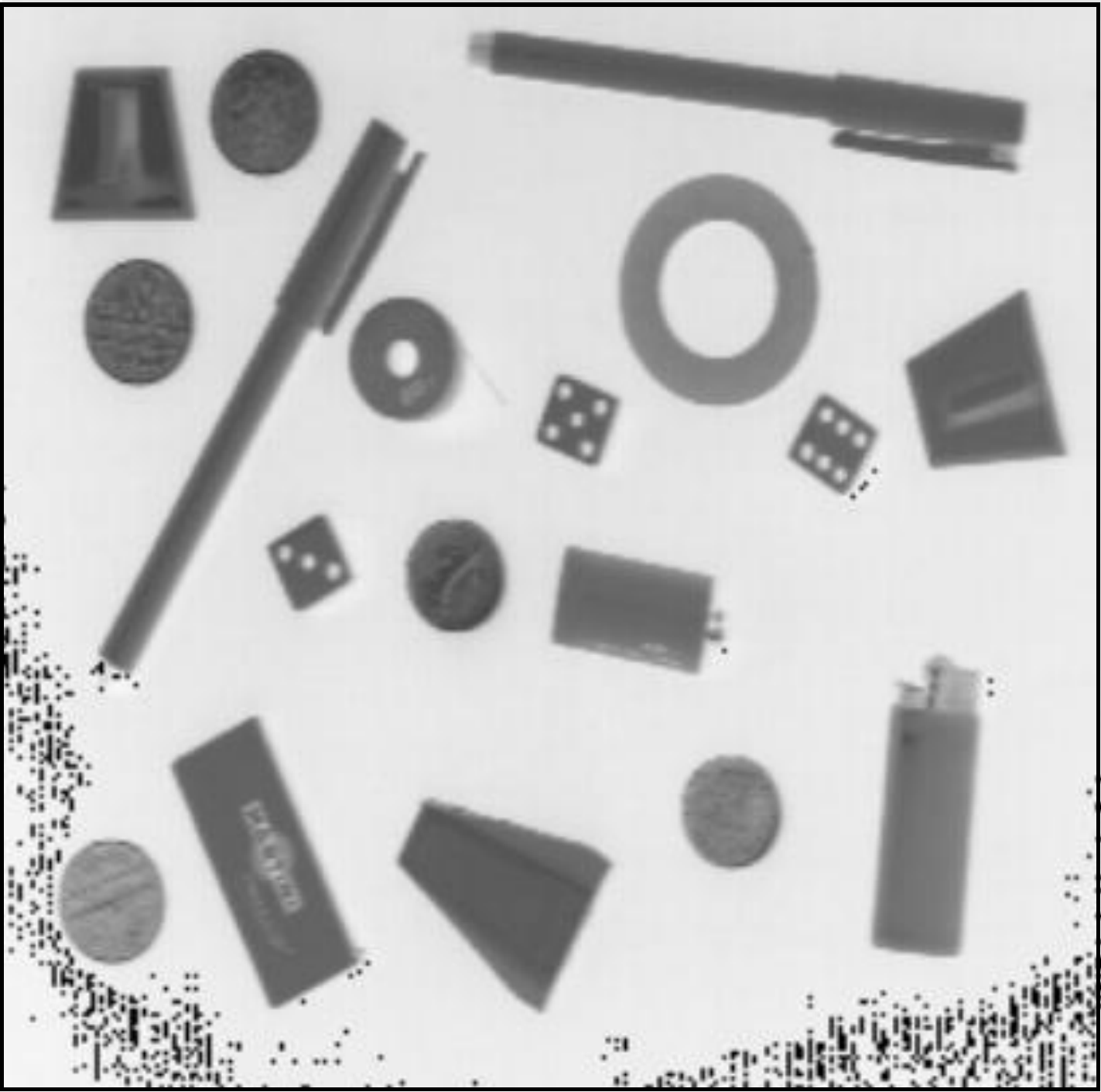}}
    \vspace*{\tinycloser}
    \centerline{(a) Input image.}
    \end{minipage}
    %
    %
    \begin{minipage}[b]{\tinympscale\linewidth}
    \centerline{\includegraphics[width=\tinyfigscale\linewidth]{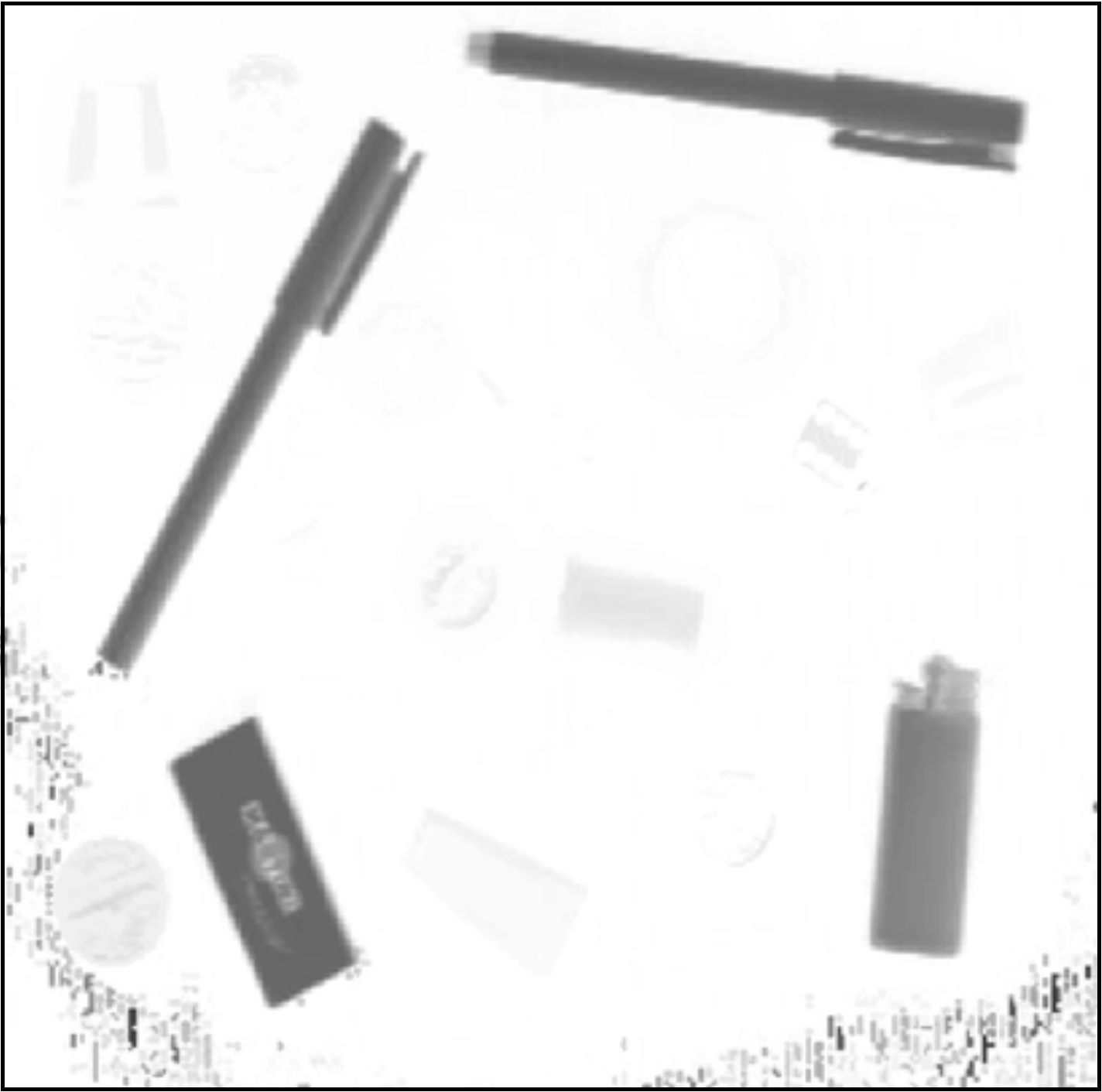}}
    \vspace*{\tinycloser}
    \centerline{(b) Shape-based leveling.}
    \end{minipage}
    \vspace*{\notsoclose}
    \newline
    %
    %
    \begin{minipage}[b]{\tinympscale\linewidth}
      \centerline{\includegraphics[width=\tinyfigscale\linewidth]{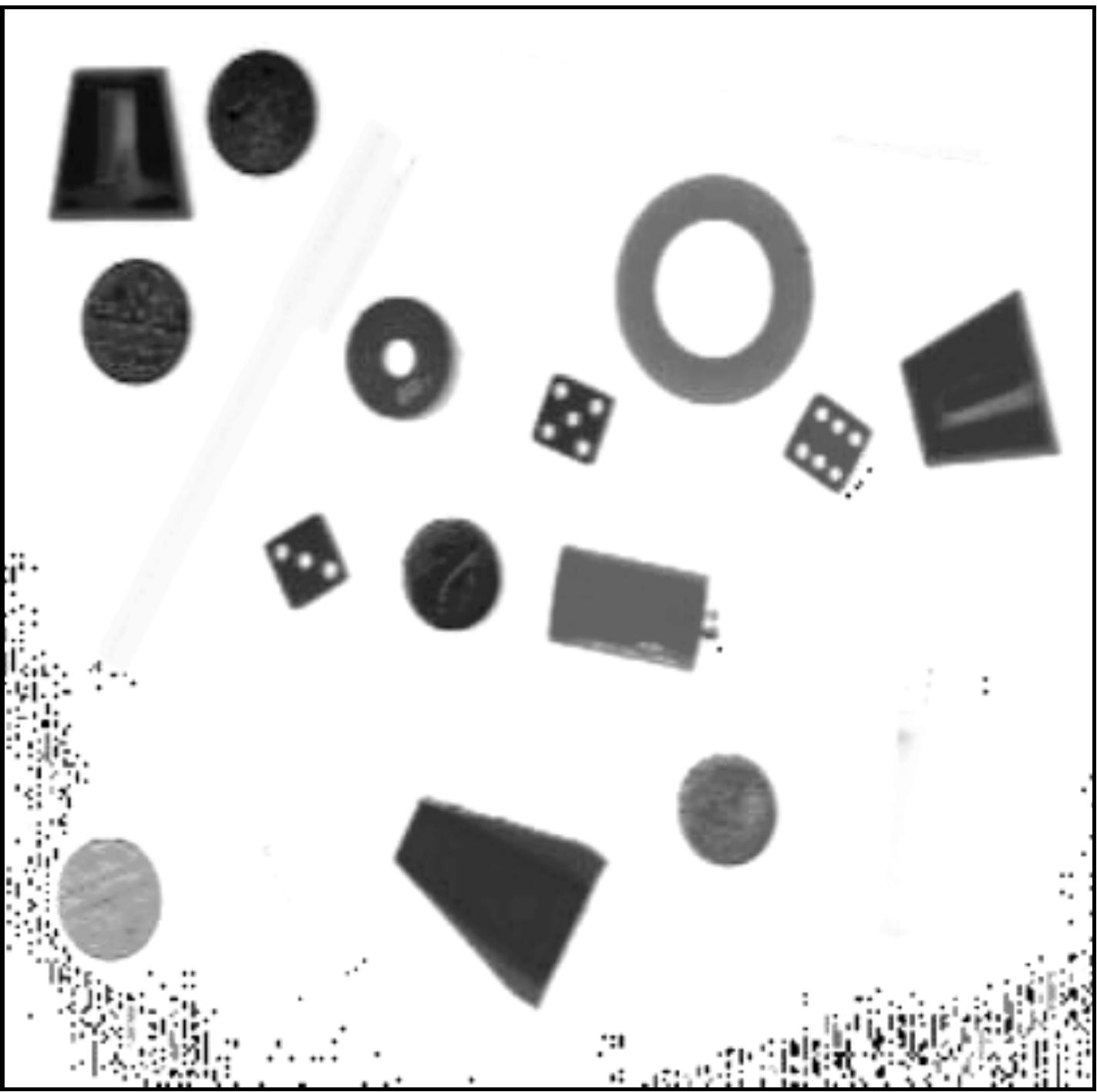}}
      \vspace*{\tinycloser}
      \centerline{(c) is (a) minus (b).}
    \end{minipage}
    %
    %
    \begin{minipage}[b]{\tinympscale\linewidth}
      \centerline{\includegraphics[width=\tinyfigscale\linewidth]{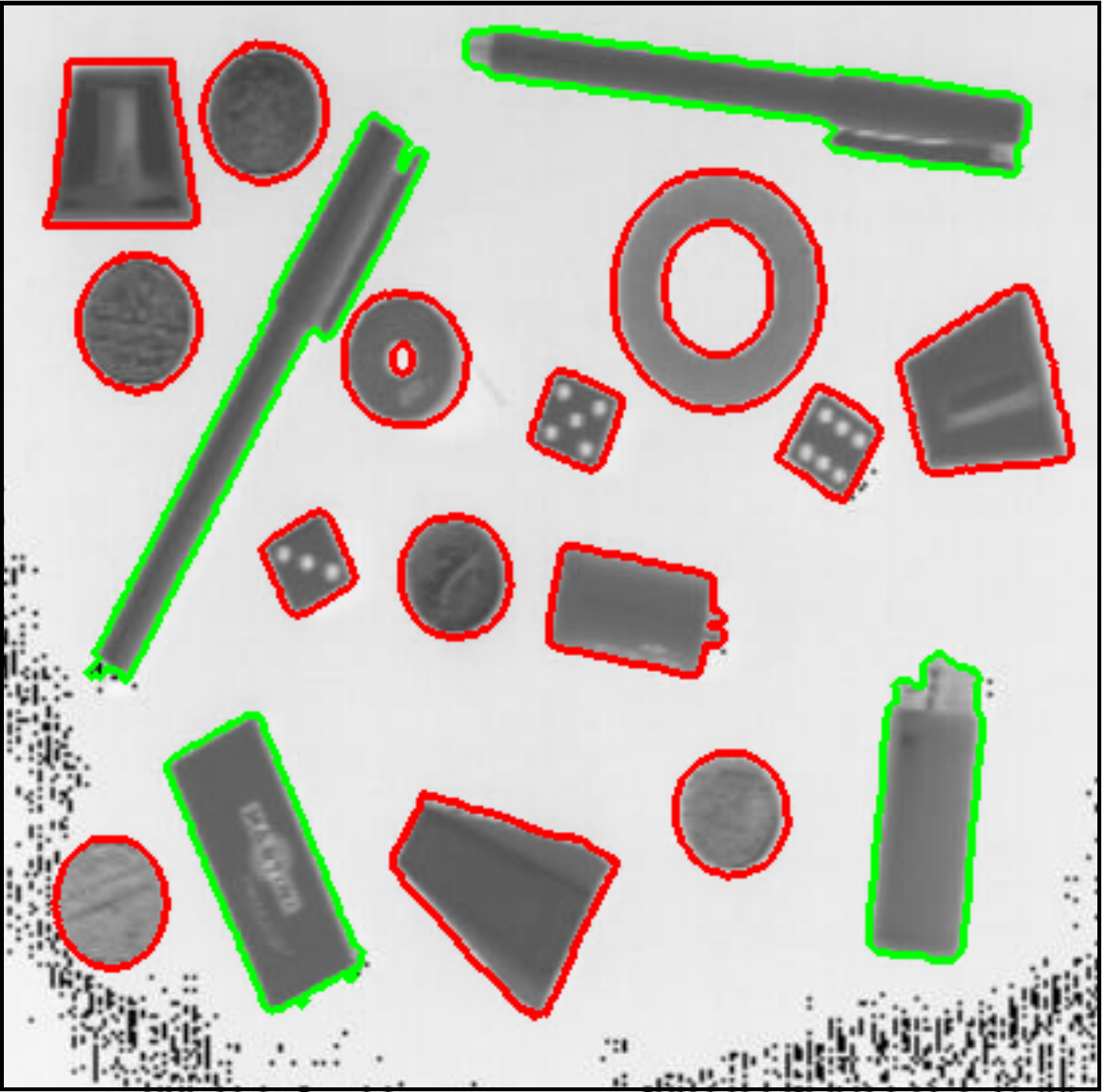}}
      \vspace*{\tinycloser}
      \centerline{(d) Object detection.}
    \end{minipage}
  \end{center}
  %
  \vspace*{-5mm}
  \caption{Illustration of shape-based leveling and object detection.}
  \label{fig:leveling}
\end{figure}


\begin{figure}
  \begin{center}
  \begin{minipage}[b]{0.8\linewidth}
    \begin{center}
      \centerline{\includegraphics[width=0.9\linewidth]{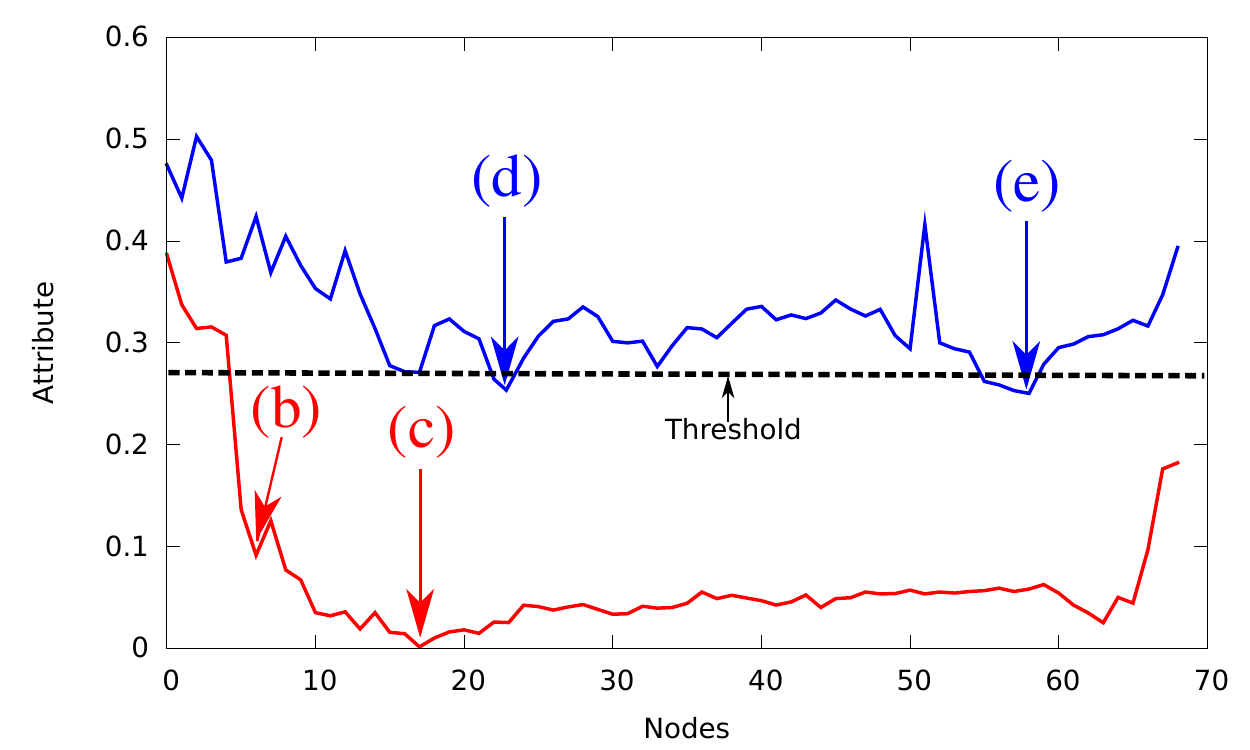}}
      \centerline{(a)}
    \end{center}
  \end{minipage}
  \vspace*{-4mm}
  \newline
  \begin{minipage}[b]{0.20\linewidth}
    \begin{center}
      \centerline{\includegraphics[width=1.0\linewidth]{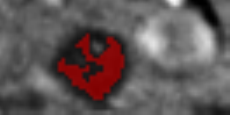}}
      \centerline{(b)}
      \vspace*{0mm}
      \centerline{\includegraphics[width=1.0\linewidth]{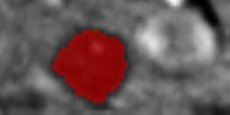}}
      \centerline{(c)}
    \end{center}
  \end{minipage}
  \begin{minipage}[b]{0.20\linewidth}
    \begin{center}
      \centerline{\includegraphics[width=1.0\linewidth]{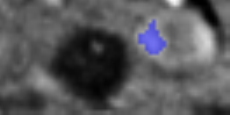}}
      \centerline{(d)}
      \vspace*{0mm}
      \centerline{\includegraphics[width=1.0\linewidth]{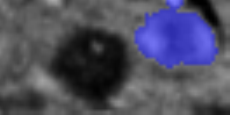}}
      \centerline{(e)}
    \end{center}
  \end{minipage}
  \begin{minipage}[b]{0.20\linewidth}
    \begin{center}
      \centerline{\includegraphics[width=1.0\linewidth]{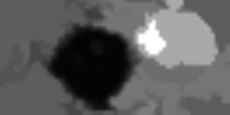}}
      \centerline{(f)}
      \vspace*{0mm}
      \centerline{\includegraphics[width=1.0\linewidth]{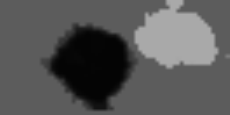}}
      \centerline{(g)}
    \end{center}
  \end{minipage}
  \vspace*{-5mm}
  \caption{(a) Evolution of ``circularity'' on two
    branches of $\Tree$; (b to e): Some shapes; (f) Attribute thresholding; (g)  Shaping.}
  \label{fig:circularity}
  \end{center}
\end{figure}

\begin{figure}
  \begin{center}
    %
    %
    \hspace*{1mm}
    \begin{minipage}[b]{\tinympscale\linewidth}
    \centerline{\includegraphics[width=\tinyfigscale\linewidth]{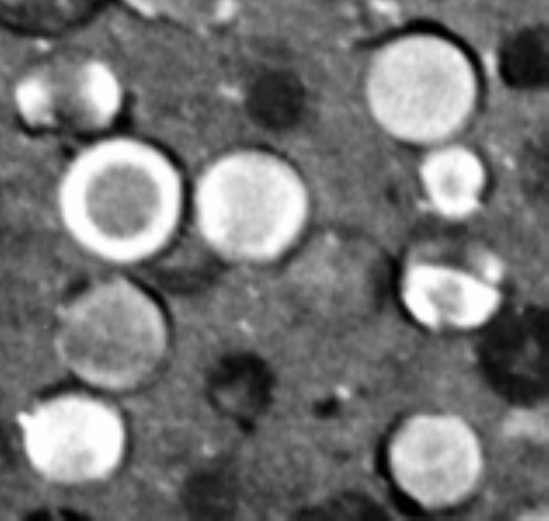}}
    \vspace*{\tinycloser}
    \centerline{(a) Input image.}
    \end{minipage}
    %
    %
    \begin{minipage}[b]{\tinympscale\linewidth}
      \centerline{\includegraphics[width=\tinyfigscale\linewidth]{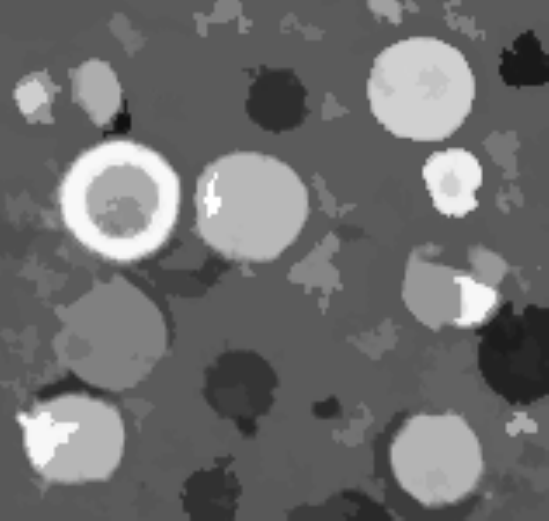}}
      \vspace*{\tinycloser}
      \centerline{(b) Shaping 1.}
    \end{minipage}
    \vspace*{\notsoclose}
    \newline
    %
    %
    \hspace*{1mm}
    \begin{minipage}[b]{\tinympscale\linewidth}
    \centerline{\includegraphics[width=\tinyfigscale\linewidth]{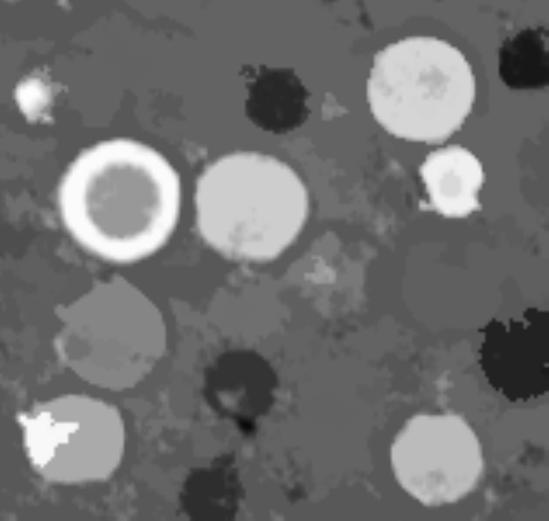}}
    \vspace*{\tinycloser}
    \centerline{(c) Low threshold of $\Attribute$.}
    \end{minipage}
    %
    %
    \begin{minipage}[b]{\tinympscale\linewidth}
      \centerline{\includegraphics[width=\tinyfigscale\linewidth]{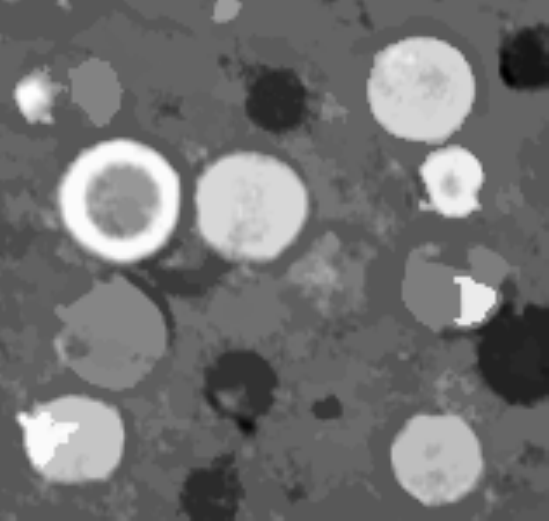}}
      \vspace*{\tinycloser}
      \centerline{(d) Higher threshold of $\Attribute$.}
    \end{minipage}
    \vspace*{\notsoclose}
    \newline
    %
    %
    ~\hspace{-1mm}
    \begin{minipage}[b]{\tinympscale\linewidth}
      \centerline{\includegraphics[width=\tinyfigscale\linewidth]{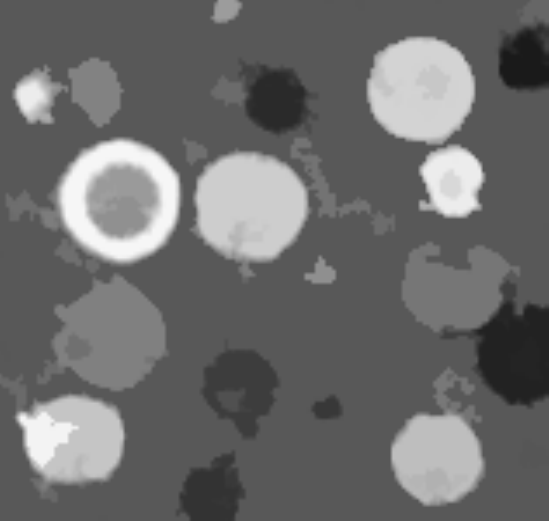}}
      \vspace*{\tinycloser} \centerline{(e) Threshold strategy.}
    \end{minipage}
    %
    %
    \begin{minipage}[b]{\tinympscale\linewidth}
      \centerline{\includegraphics[width=\tinyfigscale\linewidth]{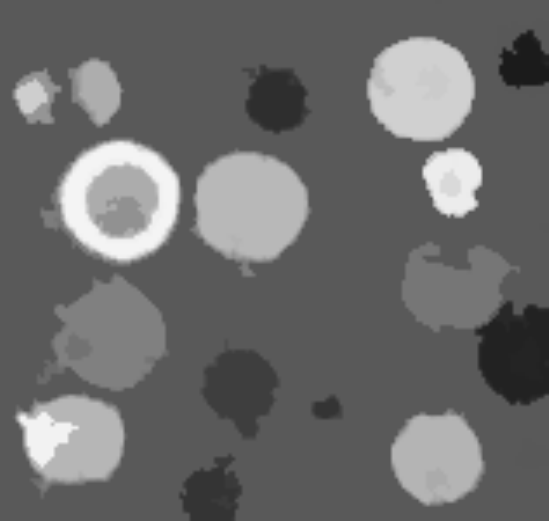}}
      \vspace*{\tinycloser}
      \centerline{(f)  Shaping 2.}
    \end{minipage}
  \end{center}
  %
  \vspace*{-5mm}
  \caption{Comparison of extinction-based shapings with
    attribute-thresholding. (b-d): Using one shape attribute; (e-f):
    Using a combination of shape attributes.}
  \label{fig:comparison}
\end{figure}

In Figure~\ref{fig:leveling}, we present a round shape-based leveling
filter.  For that, $\Tree$ is the min-tree of the input image~(a) and
$\Attribute$ is a circularity attribute.  The tree simplification
$\Tree\Tree \rightarrow \Tree\Tree'$ is performed by a morphological
closing of $\Attribute\Attribute$.  The reconstruction gives the
filtered image~(b), which is a leveling of the input image.  The
image~(c) is a top-hat (subtraction of the leveling from the input
image) that depicts the objects removed by the filter.

Figure \ref{fig:circularity} shows an example of evolution of a shape
attribute, the circularity on a simple image. The light round shape
and the dark one are both meaningful round objects compared to their
context. However, their attribute values are very different.  In order
to obtain the light one, a higher threshold is applied, but this makes
some non desired shapes appear, the shapes presented in the background
in Figure \ref{fig:circularity}~(f).

In Figure~\ref{fig:comparison}, we compare our extinction-based
self-dual shaping approach with a variant of the
state-of-the-art thresholding approach~\cite{urbach.07.pami}.  To
process both upper and lower level sets, we propose to use a tree
of shapes $\Tree$.
In this illustration $\Attribute$ is the circularity of image shapes
for (b) to (d). When the threshold of $\Attribute$ is low, we loose
some objects (see (c)).  To be able to get all expected objects, we
have to set a high threshold; however, we keep too many unwanted
objects (see (d)).  With our shaping filter, we obtain all the
expected objects as depicted in (b). The results can be improved by
combining some shape attributes. In (e) and (f), we use a combination
of circularity and the $I/A^2$~\cite{urbach.07.pami}, the moment of
inertia divided by the square of area. The combination of shape
attributes improves significantly the results. Still, our
shaping in (f) performs much better than the threshold strategy
in (e).

\SubSection{Object detection}
\label{sec:segmentation}

Some authors have proposed to rely on image representation based on
morphological trees to extract \emph{meaningful} objects.  In
~\cite{pardo.02.icip}, objects are selected when their level lines
match some criterion based on compactness and contrast.  In
\cite{cao.05.jmiv}, objects are spotted as minima of the number of
false alarms estimated on every level lines thanks to an \textit{a
  contrario} model.  We have introduced in~\cite{xu.12.icip} an energy
functional, computed on image components of the tree of shapes and
dedicated to object detection.  This functional mixes a snake-like
term and a contextual Mumford-Shah term, and its minima correspond to
potential meaningful objects.  Put differently $\Tree$ is the tree of
shapes and $\Attribute$ is the energy functional.  To extract the
significant minima (so the proper objects), we compute the min-tree
$\Tree\Tree$ of the space of shapes $\Tree$ valued with attribute
$\Attribute$; a morphological closing with attribute
$\Attribute\Attribute$ provides a simplified tree $\Tree\Tree'$; the
only minima remaining in $\Tree\Tree'$ are the relevant objects.

This strategy is valid for any general-purpose energy functional,
including any shape attribute, or a combination of shape
attributes. As such, we are able to extract objects based on their
shape features.  In Figure~\ref{fig:leveling}~(d), red and green
contours correspond to components with significant minima of
respectively a circularity attribute and an elongation attribute.

\Section{Conclusion}
\label{sec:conclusion}

This paper has presented two new classes of morphological connected
filters: new levelings (obtained with $\Tree$ being the min-tree or
the max-tree) and self-dual shapings (obtained with $\Tree$ being
the tree of shapes~\cite{monasse.00.itip}).  Those connected operators
filter image components based on some non-increasing shape criterion.
We have also shown that they encompass the usual attribute filtering
operators.  Properties of the morphological shaping (such as
conditions for idempotence) will be studied in a forthcoming extended
version. 

We have implemented those filters~\footnote{Demo available from
\url{http://olena.lrde.epita.fr/ICPR2012}.} using our C++ image
processing library~\cite{levillain.2010.icip}, available on the
Internet as free software. The present paper also demonstrates the
interest of applying image processing algorithms to different types of
graphs, in other words, the interest of genericity for image
processing.

\section*{Acknowledgments}
This work was partially supported by the Agence Nationale de la
Recherche through contract ANR-2010-BLAN-0205-03 KIDICO. The authors
would like to thank Jean Serra for fruitful discussions and for having
coined the term ``morphological shaping".

\balance
\bibliographystyle{latex12}
\bibliography{xu.2012.icpr}
\end{document}